\documentclass{article}

\usepackage[preprint]{ProbNum25}
\usepackage[T1]{fontenc}      
\usepackage[utf8]{inputenc}   
\usepackage{microtype}        
\usepackage{comment}
\usepackage{colortbl}
\usepackage{arydshln}

\usepackage{amsmath}
\usepackage{amsthm}
\usepackage{amssymb}
\usepackage{commath}
\usepackage{mathtools}
\usepackage{thmtools}
\usepackage{nicefrac}

\usepackage{cleveref}
\usepackage[hang,flushmargin]{footmisc}

\usepackage[round]{natbib}

\newcommand{\R}{\mathbb{R}} 
\newcommand{\N}{\mathbb{N}} 

\renewcommand{\b}[1]{#1} 
\newcommand{\T}{\mathsf{T}} 

\newcommand{\domain}{\Omega} 

\PassOptionsToPackage{dvipsnames}{xcolor} 

\usepackage{listings}
\usepackage[framemethod=tikz]{mdframed}
\definecolor{lightgray2}{gray}{0.9}

\probnumtitle{A Dictionary of Closed-Form Kernel Mean Embeddings}
\probnumauthors{\name{Fran\c{c}ois-Xavier Briol}\affiliation{University College London, United Kingdom}
\and\name{Alexandra Gessner}\affiliation{University of T\"ubingen, Germany}
\and\name{Toni Karvonen}\affiliation{Lappeenranta--Lahti University of Technology LUT, Lappeenranta, Finland}
\and\name{Maren Mahsereci}\affiliation{Yahoo Research, Berlin, Germany}}

\probnumabstract{Kernel mean embeddings -- integrals of a kernel with respect to a probability distribution -- are essential in Bayesian quadrature, but also widely used in other computational tools for numerical integration or for statistical inference based on the maximum mean discrepancy. These methods often require, or are enhanced by, the availability of a closed-form expression for the kernel mean embedding. However, deriving such expressions can be challenging, limiting the applicability of kernel-based techniques when practitioners do not have access to a closed-form embedding. This paper addresses this limitation by providing a comprehensive dictionary of known kernel mean embeddings, along with practical tools for deriving new embeddings from known ones.
We also provide a Python library that includes minimal implementations of the embeddings.
}

\begin{document}

\section{Introduction}
\label{sec:introduction}
\vspace{-2mm}
Let $P$ be a probability measure on a subset $\Omega$ of $\R^d$, and $K:\Omega \times \Omega \rightarrow \mathbb{R}$ be a reproducing kernel \citep{Berlinet2004} satisfying $\int_{\Omega} K(x,x) \dif P(x) < \infty$. 
This paper considers two important quantities for kernel-based methods: the \emph{kernel mean embedding} $K_P$ 
and its integral $K_{PP}$, given by
\begin{align} \label{eq:KP}
  K_P(x) & = \int_\Omega K(x, y) \dif P(y),  \\\label{eq:KPP}
  K_{PP} & = \int_\Omega \int_\Omega K(x, y) \dif P(x) \dif P(y).
\end{align}
These quantities are ubiquitous in kernel methods, but are also needed for implementing Bayesian quadrature, a probabilistic numerical method for integration which motivates this paper \citep{Hennig2022}. They are also needed for many other kernel-based numerical integration techniques, and for statistical inference. Unfortunately, $K_P$ and $K_{PP}$ are often tedious, difficult, or even impossible to compute in closed form, limiting the applicability of these algorithms. The primary purpose of this article is therefore to collect $K_P$ and $K_{PP}$ for a number of common pairs of $K$ and $P$.
We have implemented the embeddings in a Python library\footnote{\sloppy\url{https://github.com/mmahsereci/kernel_embedding_dictionary}}.

The paper is structured as follows. In \Cref{sec:applications}, we introduce the broad range of kernel-based techniques which rely on closed-form expressions for kernel mean embeddings.
 \Cref{sec:dictionary} then provides a \emph{dictionary of kernel mean embeddings} for common pairs of kernel and distribution.
 \Cref{sec:construction}  discusses common techniques for deriving new closed-form expressions for kernel mean embeddings.
 \Cref{sec:lib} introduces our Python library.
Most of the embeddings in \Cref{sec:dictionary} are not new~\citep[e.g.,][Table~1]{Briol2019}.
However, they have never been collated in one place and are currently found in disparate sources throughout the machine learning, statistics, signal processing, and numerical analysis literatures.
As a result many a hapless researcher has had to rederive these embeddings over the years.

\vspace{-2mm}
\section{Uses of Kernel Embeddings}\label{sec:applications}
\vspace{-2mm}

The following algorithms use kernel mean embeddings.

\textbf{Bayesian quadrature.} 
Numerical integration is the computational task of approximating an integral
\begin{equation}
  I(f) = \int_\Omega f(x) \dif P(x) 
\end{equation}
using evaluations of $f$ at points $X = \{ x_1, \ldots, x_n \} \subset \Omega$. A natural approach in this context is to use a quadrature rule $\hat{I}(f)=\sum_{i=1}^n w_i f(x_i)$, where each function evaluation $f(x_i)$ is assigned a weight $w_i \in \mathbb{R}$.

Bayesian quadrature~\citep{OHagan1991,Briol2019} is a probabilistic numerical method for integration. 
It typically models $f$ as a zero-mean Gaussian process $\mathrm{GP}(0, K)$ with covariance kernel $K$ that encodes prior knowledge such as differentiability, periodicity, or sparsity.
Conditioning the prior on data $\mathcal{D} = \{X,Y\}$ consisting of evaluations $Y = (f(x_1), \ldots, f(x_n))^\T$ of $f$ at some pairwise distinct nodes $X$ yields a posterior:
\begin{align}  
    I( f ) \mid \mathcal{D} \sim \mathcal{N}(\mu_\mathcal{D}, \sigma_\mathcal{D}^2) \:\: \text{, } \:\: 
    \begin{aligned}
        \mu_\mathcal{D} &= m^\T C^{-1} Y, \\ \sigma_\mathcal{D}^2 &= K_{PP} - m^\T C^{-1} m.  
    \end{aligned}
\end{align}
Here, $C = (K(x_i, x_j))_{i,j=1}^n$ is an $n \times n$ positive-definite covariance matrix, and $m = (K_P(x_1), \ldots, K_P(x_n))^\T$ a vector of kernel mean evaluations.
The name Bayesian quadrature is justified by the posterior mean being a quadrature rule with $w = (w_1,\ldots,w_n)^\T = m^\T C^{-1}$.
The computation of $K_P$ and $K_{PP}$ is a major challenge in the implementation of Bayesian quadrature.

\textbf{Integration in Kernel Spaces.} 
The quantities $K_P$ and $K_{PP}$ also play a key role in other quadrature rules. Given a reproducing kernel $K$, denote by $\mathcal{H}$ the reproducing kernel Hilbert space (RKHS) associated with $K$ and by $\lVert\cdot\rVert_{\mathcal{H}}$ the norm of $\mathcal{H}$. Assuming that $f \in \mathcal{H}$, we can bound the error of an arbitrary quadrature rule $\hat{I}(f) = \sum_{i=1}^n w_i f(x_i)$ as
\begin{align}
 | I(f) - \hat{I}(f) | \leq \|f\|_{\mathcal{H}}   \sup_{\|f\|_{\mathcal{H}} \leq 1} | I(f) - \hat{I}(f) |.
\end{align}
The second term $\text{WCE}  \coloneqq  \sup_{\|f\|_{\mathcal{H}}  \leq 1} | I(f) - \hat{I}(f) |$ is called the \emph{worst-case (integration) error} (WCE) and has the straightforward expression \cite[e.g.,][]{Briol2019}
\begin{align} \label{eq:WCE}
     \sqrt{ K_{PP} - 2 \sum_{i=1}^n w_i K_P(x_i) + \sum_{i=1}^n \sum_{j=1}^n  w_i w_j K(x_i,x_j)} .
\end{align}
The WCE can be computed when both $K_P$ and $K_{PP}$ have known expressions. The weights minimising the WCE can also be derived and are identical to the Bayesian quadrature weights. This non-Bayesian construction is called kernel quadrature \citep{Sommariva2006,Fuselier2014,Belhadji2019,Kanagawa2017,Epperly2023}.

Many other quadrature rules are constructed or analysed using the WCE and therefore require closed-form expressions for $K_P$ or $K_{PP}$. For example, quasi-Monte Carlo~\citep{Niederreiter1992,Dick2010,DickKuoSloan2013} is a set of quadrature rules for $P$ being a uniform measure. It uses equal weights and point sets selected to guarantee that the WCE decreases at a fast rate in $n$, and evaluations of the WCE are often returned as computable guarantees on the performance of the method. Relatedly, kernel herding \citep{Chen2010,Bach2012,Lacoste-julien2015} also uses equal weights, but selects points by directly minimising the WCE by repeatedly evaluating it. 

Beyond these, many algorithms aim to minimise the WCE in $\mathcal{H}$ but rely on sample-based approximations of $K_P$ and/or $K_{PP}$. These algorithms are typically designed this way so as to make them more widely applicable, but they would most likely benefit from access to closed-form expressions. Examples include gradient flows (\citealp{Arbel2019,Hertrich2024,Chen2024deregularizedmmd}; see also \citealp{Xu2022} and \citealp{Belhadji2025} for the benefits of closed-forms), thinning algorithms \citep{Dwivedi2024}, and quadrature rules based on leverage scores \citep{Chatalic2023}.

\textbf{Statistical Inference with Kernel Embeddings.}
The maximum mean discrepancy (MMD; \citealp{Gretton2012}) is a probability metric under mild condition on the kernel \citep{Sriperumbudur2009}. It compares two distribution by the magnitude of the difference in their mean embedding measured in the RKHS norm. The MMD admits a straightforward expression:
\begin{equation}
\begin{split}
    \text{MMD}^2(P,Q) 
    & \coloneqq \| K_P - K_Q \|_{\mathcal{H}}^2 \\
    & = K_{PP} - 2 K_{PQ} + K_{QQ}
    \end{split}
\end{equation}
where $K_{PQ} = \int_\Omega \int_\Omega K(x, y) \dif P(x) \dif Q(y)$. Note that when $Q$ is the empirical measure $Q_n \coloneqq  \sum_{i=1}^n w_n \delta_{x_i}$, the MMD becomes exactly the WCE from \Cref{eq:WCE}. 

The convenient expression of the MMD has led to multitudes of applications of kernel mean embeddings \citep{Muandet2016}. For example,  goodness-of-fit testing can be performed using $\text{MMD}^2(P,Q_n)$ as a test statistic \citep{Lloyd2015,Kellner2019}, where $P$ is the model under the null hypothesis and $Q_n$ the observed data. Another example is parametric estimation through minimum distance estimation \citep{Briol2019estimator,Cherief-Abdellatif2019,Alquier2023}, approximate Bayesian computation, generalised Bayesian inference \citep{Cherief-Abdellatif2020_MMDBayes,Dellaporta2022,Pacchiardi2024}, or variational inference \citep{Huang2023}, which typically make use of evaluations of $\text{MMD}^2(P_\theta,Q_n)$, where $P_\theta$ is the parametric model and $Q_n$ the observed data.

 Many other algorithms in statistics and machine learning either have limited applicability due to the lack of tractability of $K_P$ and $K_{PP}$, or require that these quantities are approximated through samples, which introduces additional error \citep{Chamakh2024}. This includes two-sample testing \citep{Gretton2012}, causal inference \citep{Singh2019,Singh2024,Muandet2021,Sejdinovic2024}, density estimation \citep{Song2008}, the analysis of variance \citep{Durrande2013}, linked emulation \citep{MingGuillas2020}, kernel Bayes  \citep{Fukumizu2013}, auto-encoders \citep{Tolstikhin2018,Rustamov2021} and generative adversarial networks \citep{Dziugaite2015}.

  \vspace{-2mm}

\section{Dictionary of Embeddings}
\label{sec:dictionary}

  \vspace{-2mm}

Now that we have highlighted the importance of $K_P$ and $K_{PP}$, this section collates expressions for some commonly used kernels and distributions. We focus primarily on the uniform and Gaussian, though other formulae exist in the literature~\citep[e.g.,][]{Nishiyama2016,Nishiyama2020}.
Our expressions are summarised in Table~\ref{kmes}.
In absence of references we only sketch the derivations, which are in most cases tedious and uninteresting.
Our Python library and its tests can be thought of as ``numerical proofs'' of the identities here.

  \textbf{Uniform Distribution.} Let $a_i < b_i$ for $i = 1,\ldots,d$ and define $r_i = b_i - a_i > 0$. The \emph{uniform measure} on $\domain = \lbrack a_1, b_1 \rbrack \times \cdots \times \lbrack a_d, b_d \rbrack \subset \R^d$ has density
  \begin{equation} \label{eq:lebesgue-measure}
    p(\b{x}) = \prod_{i=1}^d (b_i - a_i)^{-1} = \prod_{i=1}^d r_i^{-1}.
  \end{equation}

  \textbf{Gaussian Distribution.} Let $\b{\Sigma} \in \R^{d \times d}$ be a positive-definite matrix and $\b{\mu} \in \R^d$. A \emph{Gaussian measure} on $\domain = \R^d$ has the density
\begin{equation} \label{eq:gaussian-measure}
  p(\b{x}) = C(d, \Sigma) \cdot \exp\bigg( \! - \frac{1}{2} (\b{x} - \b{\mu})^\T \b{\Sigma}^{-1} (\b{x} - \b{\mu}) \bigg),
\end{equation}
where $C(d, \Sigma) = (2\pi)^{-d/2} (\det \b{\Sigma})^{-1/2}$.
The \emph{centered} Gaussian distribution has $\b{\mu}=0$.
The \emph{isotropic Gaussian distribution} has a diagonal covariance $\b{\Sigma} = \sigma^2 \b{I}$ for $\sigma > 0$.

\begin{table*}[thb]\centering
\caption{References to the equations for kernel mean embeddings of kernel-measure pairs.}
    \begin{tabular}{lclccc}
        \toprule
        Kernel   & Degree
         & Measure $P$  & Domain $\domain$ &   $K_P$ &  $K_{PP}$  \\
        \midrule
        Gaussian && Uniform & $\prod_{i=1}^{d} [a_i, b_i]$ &   \eqref{eq:Kp_gauss_uniform} & \eqref{eq:Kpp_gauss_uniform} \\
         && Gaussian & $\R^d$ &   \eqref{eq:Kp_gauss_gauss}  & \eqref{eq:Kpp_gauss_gauss}\\
        \cmidrule(l){1-6}
        Mat\'ern & $n + \nicefrac{1}{2}$ & Uniform & $[a,b]$ &  \eqref{eq:matern-ginsbourger} & \eqref{eq:Kpp_matern_general_uniform} \\
        & \nicefrac{1}{2}, \nicefrac{3}{2}, \nicefrac{5}{2}, \nicefrac{7}{2} & Uniform & $[a,b]$ &  \eqref{eq:Kp_matern12_uniform}, \eqref{eq:Kp_matern32_uniform}, \eqref{eq:Kp_matern52_uniform}, \eqref{eq:Kp_matern72_uniform}  & \eqref{eq:Kpp_matern12-52_uniform}, \eqref{eq:Kpp_matern32-72_uniform} \\
         & \nicefrac{1}{2}, \nicefrac{3}{2}, \nicefrac{5}{2} & Gaussian & $\R$ &   \eqref{eq:Kp_matern12_gauss}, \eqref{eq:Kp_matern32_gauss}, \eqref{eq:Kp_matern52_gauss}  & ?, ?, ?  \\  
        \cmidrule(l){1-6}
        Wendland & 0 & Uniform & $[a,b]$ &   \eqref{eq:Kp_wendland0_uniform}  & \eqref{eq:Kpp_wendland0_uniform}\\
         & 0, 2 & Gaussian & $\R$ &   \eqref{eq:Kp_wendland0_gauss}, \eqref{eq:Kp_wendland2_gauss}  & ?, ? \\
        \cmidrule(l){1-6}
        Brownian Motion & $H$ & Uniform & $[a,b]$ & \eqref{eq:Kp_fbm_uniform}  & \eqref{eq:Kpp_fbm_uniform}\\
        \cmidrule(l){1-6}
        Power series & & Uniform & $\prod_{i=1}^{d} \lbrack a_i, b_i\rbrack$ & \eqref{eq:Kp_powerseries_uniform} & \eqref{eq:Kpp_powerseries_uniform}\\
         & & Gaussian & $\R^d$ & \eqref{eq:Kp_powerseries_gaussian} & \eqref{eq:Kpp_powerseries_gaussian}\\
        \cmidrule(l){1-6}
        Sobolev & $\nicefrac{3}{2}$& Uniform & $\mathbb{S}^2$ & \eqref{kp_kpp_sobolev32_sphere} & \eqref{kp_kpp_sobolev32_sphere}\\
         & $\infty$ & Uniform & $\mathbb{S}^2$ & \eqref{kp_kpp_sobolevinfty_sphere} & \eqref{kp_kpp_sobolevinfty_sphere}\\
        \cmidrule(l){1-6}
        Periodic Sobolev & 2r & Uniform & $\mathbb{S}^1$ or $\lbrack 0,1 \rbrack$ & \eqref{eq:korobov-embeddings} & \eqref{eq:korobov-embeddings} \\
        \cmidrule(l){1-6}
        Stein & & Unnormalised & $\mathbb{R}^d$ & \eqref{eq:stein_embeddings} & \eqref{eq:stein_embeddings}\\
        \bottomrule
    \end{tabular}
    \label{kmes}
\end{table*}

\vspace{-2mm}
\noindent\parbox{\textwidth}{
  \vspace{-0.7cm}
  \subsection{Gaussian Kernel}
  Let $\b{\Lambda} \in \R^{d \times d}$ be a positive-definite length-scale matrix. The \emph{Gaussian kernel} is
  \begin{equation}
    K(\b{x}, \b{y}) = \exp\bigg( -\! \frac{1}{2} (\b{x} - \b{y})^\T \b{\Lambda}^{-1} (\b{x} - \b{y}) \bigg) \quad \text{ for } \quad x, y \in \R^d.
  \end{equation}

  \textbf{Uniform Distribution.} Consider the uniform distribution in~\eqref{eq:lebesgue-measure} and suppose that the length-scale matrix is $\Lambda = \operatorname{diag}(\ell_1^2, \ldots, \ell_d^2)$ for $\ell_1, \ldots, \ell_d > 0$.
  Then
  \begin{align}
    K_P(\b{x}) &= \bigg( \frac{\pi}{2} \bigg)^{d/2}\prod_{i=1}^d \frac{\ell_i }{r_i} \Bigg\lbrack \operatorname{erf}\bigg( \frac{b_i-x_i}{\ell_i\sqrt{2}}\bigg) - \operatorname{erf}\bigg(\frac{a_i-x_i}{\ell_i\sqrt{2}}\bigg) \Bigg\rbrack , \label{eq:Kp_gauss_uniform} \\    
    K_{PP} &=( 2 \pi )^{d/2}\prod_{i=1}^d \frac{\ell_i }{r_i^2} \Bigg\lbrack \frac{\ell_i\sqrt{2} }{\sqrt{\pi}}\bigg( \exp\bigg(\!-\frac{r_i^2}{2\ell_i^2} \bigg) - 1 \bigg) + r_i \operatorname{erf}\bigg( \frac{r_i}{\ell\sqrt{2}} \bigg) \Bigg\rbrack,
      \label{eq:Kpp_gauss_uniform}
  \end{align}
  where $\operatorname{erf}(x) = (\pi)^{-1/2} \int_{-x}^x \exp(-t^2) \dif t$ is the error function.
  It is straightforward to derive~\eqref{eq:Kp_gauss_uniform} and~\eqref{eq:Kpp_gauss_uniform} from the definition of the error function.
  We are not aware of closed-form expressions for non-diagonal lengthscale matrices.

\vspace{0.2cm}
\textbf{Gaussian Distribution.} Consider the Gaussian distribution in~\eqref{eq:gaussian-measure}.
  Then
  \begin{align}
    K_P(\b{x}) &= \det(\b{I} + \b{\Sigma} \b{\Lambda}^{-1})^{-1/2} \exp\bigg(\!-\frac{1}{2}(\b{x}-\b{\mu})^\T (\b{\Lambda} + \b{\Sigma})^{-1} (\b{x}-\b{\mu}) \bigg) ,
    \label{eq:Kp_gauss_gauss} 
    \\
    K_{PP} &= \det(\b{I} + \b{\Sigma} \b{\Lambda}^{-1})^{-1/2} \det(\b{I} + \b{\Sigma} (\b{\Lambda} + \b{\Sigma})^{-1})^{-1/2} = \sqrt{\frac{\det(\b{\Lambda})}{\det(\b{\Lambda} + 2\b{\Sigma})}}.
    \label{eq:Kpp_gauss_gauss}
  \end{align}
If $\b{\Lambda}$ and $\b{\Sigma}$ are diagonal with diagonal elements $\ell^2_i$ and $\sigma^2_i$, respectively, the expressions simplify to
  \begin{equation}
    K_P(\b{x}) = \prod_{i=1}^d \sqrt{\frac{\ell^2_i}{\ell^2_i + \sigma^2_i}} \exp\bigg(\!- \frac{(x_i - \mu_i)^2}{2(\ell^2_i + \sigma^2_i)} \bigg)
    \quad \text{ and } \quad
    K_{PP} = \prod_{i=1}^d \sqrt{\frac{\ell^2_i}{\ell^2_i + 2\sigma^2_i}}.
  \end{equation}
  Equations~\eqref{eq:Kp_gauss_gauss} and~\eqref{eq:Kpp_gauss_gauss} are derived from the usual completion of the square trick.
  The embedding for log-Gaussian distributions can be obtained by using a log-transformation on the Gaussian kernel as per \cite{Chen2024conditionalbq}.

  \subsection{Matérn Kernels}
    Let $\ell$ be a positive length-scale parameter. The \emph{Matérn kernel} of \emph{order} $\nu > 0$ is
  \begin{equation} \label{eq:matern-kernel}
    K^\nu(\b{x}, \b{y}) = \frac{2^{1-\nu}}{\Gamma(\nu)} \big( \sqrt{2\nu} \, \tau \big)^\nu \mathrm{K}_\nu \big( \sqrt{2\nu} \, \tau \big), \quad \text{ with } \quad \tau = \frac{\lVert x - y \rVert_2}{\ell} \quad \text{and} \quad x, y \in \R^d,
  \end{equation}
  where $\mathrm{K}_\nu$ is the modified Bessel function of the second kind of order $\nu$ and $\Gamma$ is the gamma function.
}

\twocolumn[
For $\nu = n + 1/2$, $n \in \N_0$, the kernel has a more elementary form:
\begin{equation}
  K^{n+1/2}(\b{x}, \b{y}) = \exp\big( -\sqrt{2n+1} \, \tau \big) \frac{n!}{(2n)!} \sum_{k=0}^n \frac{(n+k)!}{k!(n-k)!} \big( 2\sqrt{2n+1} \, \tau \big)^{n-k}.
\end{equation}

For $n \in \{ 0,1,2,3 \}$ these are
\begin{alignat}{3}
  &K^{1/2}(\b{x}, \b{y}) = \exp(- \tau ), \quad\quad 
  &&K^{5/2}(\b{x}, \b{y}) = \bigg(1 + \sqrt{5}\, \tau + \frac{5}{3} \tau^2 \bigg) \exp\big(\!-\sqrt{5}\,\tau\big), \\
  &K^{3/2}(\b{x}, \b{y}) = \big(1+\sqrt{3}\, \tau \big) \exp\big(\!-\sqrt{3}\, \tau \big), \quad\quad 
  &&K^{7/2}(\b{x}, \b{y}) = \bigg(1 + \sqrt{7}\, \tau + \frac{14}{5} \tau^2 + \frac{7^{3/2}}{15} \tau^3 \bigg) \exp\big(\!- \sqrt{7}\, \tau \big) .
\end{alignat}

\textbf{Uniform Distribution ($d=1$).} Consider the  uniform distribution in~\eqref{eq:lebesgue-measure} on $\domain = \lbrack a,b \rbrack \subset \R$ with $a < b$.
Let $r = b - a > 0$. Then 
\begin{equation} \label{eq:matern-ginsbourger}
  K_P^{n + 1/2}(x) = \frac{\alpha_n}{r} \cdot \frac{n!}{(2n)!} \bigg\lbrack 2 c_{n,0} - Q_n\bigg(\frac{x - a}{\alpha_n}\bigg) - Q_n\bigg( \frac{b - x}{\alpha_n}\bigg)\bigg\rbrack ,
\end{equation}
where
\begin{equation}
  \alpha_n = \frac{\ell}{\sqrt{2n+1}}, \quad c_{n, m} = \frac{1}{m!} \sum_{i=0}^{n-m} \frac{(n+i)!}{i!} 2^{n-i}, \quad \text{ and } \quad Q_n(z) = e^{-z} \sum_{m=0}^n c_{n,m} z^m .
\end{equation}
This formula is obtained from a formula for $\Omega = \lbrack 0, 1 \rbrack$ in Section~9 of \citet{Ginsbourger2016} with a change of variables.
From an additional change of variables and the formula $\gamma(m+1, x) = \int_0^x t^m e^{-t} \dif = m! (1 - e^{-x} \sum_{i=0}^m x^i / i!)$ for the lower incomplete gamma function at $m \in \N_0$ it follows that
\begin{equation} \label{eq:Kpp_matern_general_uniform}
  K_{PP}^{n + 1/2} = \frac{2 \alpha_n^2}{r^2} \cdot \frac{n!}{(2n)!} \bigg\lbrack \frac{r}{\alpha_n} c_{n,0} - \sum_{m=0}^n c_{n,m} \gamma_{m+1} \bigg\rbrack, 
  \quad \text{ where } \quad 
  \gamma_{m+1} = m! \bigg\lbrack 1 - \exp\bigg(\!-\frac{r}{\alpha_n}\bigg) \sum_{i=0}^m \frac{1}{i!} \bigg( \frac{r}{\alpha_n}\bigg)^i \, \bigg \rbrack .
\end{equation}
Let $d_n (x, y) = (x-y)/\alpha_n$ and $\rho_n = r/\alpha_n$.
Then, for $n \in \{0, 1, 2, 3\}$ the embeddings are 
\begin{align}
  K_P^{1/2}(x) ={}& \frac{1}{\rho_0} \big\lbrack 2 - \exp\big(d_0 (a, x)\big) - \exp\big( d_0 (x, b)\big) \big\rbrack, \label{eq:Kp_matern12_uniform} \\
  K_P^{3/2}(x) ={}& \frac{1}{\rho_1}  \big\lbrack 4 - \exp\big(d_1 (x, b)\big) \big(2 - d_1 (x, b)\big) - \exp\big(d_1 (a, x)\big)\big(2 - d_1 (a, x)\big)\big\rbrack, \label{eq:Kp_matern32_uniform} \\
  \begin{split}
  K_P^{5/2}(x) ={}& \frac{1}{3 \rho_2}  \big\lbrack 16 - \exp\big(d_2 (x, b)\big) \big(8 - 5 d_2 (x, b) + d_2 (x, b)^2 \big)
    - \exp\big(d_2 (a, x)\big) \big(8 - 5 d_2 (a, x)  + d_2 (a, x)^2  \big) \big\rbrack,
    \end{split}\label{eq:Kp_matern52_uniform} 
    \\
  \begin{split}
  K_P^{7/2}(x) ={}& \frac{1}{15 \rho_3}  \big\lbrack 96- \exp\big(d_3 (x, b)\big) \big( 48 - 33 d_3 (x, b) + 9 d_3 (x, b)^2 - d_3 (x, b)^3\big) \bigg. \\
    &\hspace{1.2cm} \bigg. - \exp\big(d_3 (a, x)\big) \big( 48 - 33 d_3 (a, x) + 9 d_3 (a, x)^2 - d_3 (a, x)^3\big) \big\rbrack. \label{eq:Kp_matern72_uniform} 
    \end{split}
\end{align}
The integrals of these four embeddings are
\begin{alignat}{3}
  &K_{PP}^{1/2} = \frac{2}{\rho_0^2} \big\lbrack \rho_0 - 1 + \exp(-\rho_0 )\big\rbrack, \quad
  &&K_{PP}^{5/2} = \frac{2}{3 \rho_2^2} \big\lbrack 8 \rho_2 - 15 + \exp(-\rho_2) ( \rho_2^2 + 7 \rho_2 + 15) \big\rbrack, \label{eq:Kpp_matern12-52_uniform} \\
  &K_{PP}^{3/2} = \frac{2}{\rho_1^2} \big\lbrack 2\rho_1 - 3 + \exp(-\rho_1 ) (\rho_1 + 3 ) \big\rbrack, \quad
  &&K_{PP}^{7/2} = \frac{2}{15 \rho_3^2} \big\lbrack 3 (16 \rho_3 - 35 ) + \exp(-\rho_3 ) ( \rho_3^3 + 12 \rho_3^2 + 57 \rho_3 +105 ) \big\rbrack. \label{eq:Kpp_matern32-72_uniform}
\end{alignat}

\textbf{Gaussian Distribution ($d=1$).}
Consider the centered univariate Gaussian distribution in~\eqref{eq:gaussian-measure} with $\Sigma = \sigma^2$ and let $\Phi(x) = \frac{1}{2} \lbrack 1 + \mathrm{erf}(x/\sqrt{2}) \rbrack$ denote the cumulative distribution function of the standard normal distribution. \citet{MingGuillas2020} have derived the following three mean embeddings:
\begin{align}
    K_{P}^{1/2}(x) ={}& \exp\bigg( \frac{\sigma^2 + 2 \ell (x - \mu)}{2\ell^2} \bigg) \Phi\bigg( \frac{\mu - \sigma^2/\ell - x}{\sigma} \bigg) + \exp\bigg( \frac{\sigma^2 - 2 \ell (x - \mu)}{2\ell^2} \bigg) \Phi\bigg( \frac{x - \mu - \sigma^2/\ell}{\sigma} \bigg) \label{eq:Kp_matern12_gauss}\\
    \begin{split}
    K_{P}^{3/2}(x) ={}& \exp\bigg( \frac{3\sigma^2 + 2\sqrt{3} \ell (x - \mu)}{2\ell^2} \bigg) \bigg\lbrack \bigg( 1 - \frac{\sqrt{3} (x-\mu_1)}{\ell}\bigg) \Phi\bigg( \frac{\mu_1 - x}{\sigma} \bigg) + \sqrt{\frac{3 \sigma^2 }{2\pi \ell^2}} \exp\bigg(\!-\frac{(\mu_1 - x)^2}{2\sigma^2}\bigg) \bigg\rbrack \\
    &+ \exp\bigg( \frac{3\sigma^2 - 2\sqrt{3} \ell (x - \mu)}{2\ell^2} \bigg) \bigg\lbrack \bigg( 1 + \frac{\sqrt{3} (x - \mu_2)}{\ell}\bigg) \Phi\bigg( \frac{x - \mu_2}{\sigma} \bigg) + \sqrt{\frac{3 \sigma^2 }{2\pi \ell^2}} \exp\bigg(\!-\frac{(x - \mu_2)^2}{2\sigma^2}\bigg) \bigg\rbrack ,
    \end{split} \label{eq:Kp_matern32_gauss}
\end{align}
]

\twocolumn[
where $\mu_1 = \mu - \sqrt{3} \sigma^2 / \ell$ and $\mu_2 = \mu + \sqrt{3} \sigma^2 / \ell$, and
\begin{equation}
    \begin{split}
        K_P^{5/2}(x) ={}& \exp\bigg( \frac{5\sigma^2 + 2\sqrt{5} \ell (x - \mu)}{2\ell^2} \bigg) \bigg\lbrack \bigg( 1 - \frac{\sqrt{5} (x-\mu_3)}{\ell} + \frac{5(x^2 - 2\mu_3 x + \mu_3^2 + \sigma^2)}{3\ell^2} \bigg) \Phi\bigg( \frac{\mu_3 - x}{\sigma} \bigg) \bigg. \\
        &\hspace{6.5cm} + \bigg. \bigg(\frac{\sqrt{5}}{\ell} + \frac{5(\mu_3 - x)}{3\ell^2} \bigg) \frac{\sigma}{\sqrt{2\pi}} \exp\bigg(\!-\frac{(\mu_3 - x)^2}{2\sigma^2}\bigg) \bigg\rbrack \\
    &+ \exp\bigg( \frac{5\sigma^2 - 2\sqrt{5} \ell (x - \mu)}{2\ell^2} \bigg) \bigg\lbrack \bigg( 1 + \frac{\sqrt{5} (x-\mu_4)}{\ell} + \frac{5(x^2 - 2\mu_4 x + \mu_4^2 + \sigma^2)}{3\ell^2} \bigg) \Phi\bigg( \frac{x - \mu_4}{\sigma} \bigg) \bigg. \\
        &\hspace{6.5cm} + \bigg. \bigg(\frac{\sqrt{5}}{\ell} + \frac{5(x - \mu_4))}{3\ell^2} \bigg) \frac{\sigma}{\sqrt{2\pi}} \exp\bigg(\!-\frac{(x - \mu_4)^2}{2\sigma^2}\bigg) \bigg\rbrack ,
    \end{split} \label{eq:Kp_matern52_gauss}
\end{equation}
where $\mu_3 = \mu - \sqrt{5} \sigma^2 / \ell$ and $\mu_4 = \mu + \sqrt{5} \sigma^2 / \ell$.
See \cite{MingGuillas2020} for more related formulae.

\subsection{Wendland Kernels}
\emph{Wendland kernels} \citep{Wendland1995} are compactly supported kernels on $\R^d$ and give rise to sparse Gram matrices.
Let $\ell$ be a positive length-scale parameter.
The first three even-order Wendland kernels are given by
\begin{align}
  K^0(x, y) = (1-\tau)_{+}, \quad K^2(x, y) = (1-\tau)^3_{+} (3\tau+1) \quad \text{and} \quad K^4(x, y) = (1-\tau)^5_{+} (8\tau^2 + 5\tau+1) \quad 
\end{align}
for $x, y \in \R^d$, where $\tau = \lVert x - y \rVert / \ell$ and $(a)_{+} \coloneqq \max (0,a)$.
Because the kernel is defined piecewise, the mean embeddings are exceptionally unwieldy. We only include selected embeddings for kernels of orders zero and two.
We have computed these embeddings with Mathematica.

\vspace{0.2cm}
\textbf{Uniform Distribution ($d=1$).}
Consider the uniform distribution in~\eqref{eq:lebesgue-measure} on $\Omega = \lbrack a, b \rbrack \subset \R$ with $a < b$.
Denote $r = b - a > 0$.
Then
\begin{equation}
  K_P^0(x) = 
  \begin{dcases}
    \tfrac{\ell}{r} &\quad \text{ if } \quad b \geq x + \ell \text{ and } a + \ell < x, \\
    \tfrac{1}{2r\ell} \lbrack 2 x(a  + \ell ) + \ell^2 - a^2  - 2 a \ell - x^2 \rbrack &\quad \text{ if } \quad b \geq x + \ell \text{ and } a + \ell \geq x, \\
    \tfrac{1}{2r\ell} \lbrack 2b ( \ell + x) + \ell^2 - b^2 - 2 \ell x - x^2 \rbrack &\quad \text{ if } \quad b < x + \ell \text{ and } a + \ell < x, \\
    \tfrac{1}{2r\ell} \lbrack 2(b\ell + bx + ax) - a^2  - b^2 - 2(a\ell + x^2) \rbrack &\quad \text{ otherwise}
  \end{dcases} \label{eq:Kp_wendland0_uniform}
\end{equation}
and
\begin{equation}
  K_{PP}^0 = 
  \begin{dcases}
    \tfrac{5}{12} &\quad \text{ if } \quad r = 2\ell, \\
    \tfrac{1}{3r^2} \ell(3r - \ell) &\quad \text{ if } \quad \ell < r \text{ and } r \neq 2\ell, \\
    1  - \tfrac{1}{3 \ell} r &\quad \text{ if } \quad r > \ell, \\
    \tfrac{1}{3r^2} \ell(9r - 2\ell) + \tfrac{1}{3\ell} r - 2 &\quad \text{ otherwise.}
  \end{dcases} \label{eq:Kpp_wendland0_uniform}
\end{equation}
Similar expressions are available for Wendland kernels of higher order, but these are omitted due to their complexity.

\vspace{0.2cm}
\textbf{Gaussian Distribution ($d=1$).}
Consider the centered univariate Gaussian distribution in~\eqref{eq:gaussian-measure} with $\Sigma = \sigma^2$ and let \smash{$\operatorname{erf}(x) = (\pi)^{-1/2} \int_{-x}^x \exp(-t^2) \dif t$} again denote the standard error function, $\varphi(x) = \exp(-x^2/(2\sigma^2))$ the unnormalised Gaussian density function, and $s = \sqrt{2} \sigma$.
Then
\newcommand{\qqqq}{\hspace{0.7cm}}
\begin{align}
  K_P^0(x) ={}& 
  \frac{1}{2\ell} \bigg\lbrack (\ell-x) \operatorname{erf} \bigg(\frac{\ell-x}{s}\bigg)+(\ell+x) \operatorname{erf} \bigg(\frac{\ell + x}{s}\bigg)-2 x \operatorname{erf} \bigg(\frac{x}{s}\bigg)+ \frac{s}{\sqrt{\pi }} \lbrack \varphi(\ell - x) + \varphi(\ell + x) - 2\varphi(x) \rbrack \bigg\rbrack, \label{eq:Kp_wendland0_gauss}\\
  \begin{split} 
  K_P^{2}(x) 
  ={}& \frac{1}{2\ell^4} \Bigg\lbrack
    \frac{s}{\sqrt{\pi }}  \bigg\lbrack \bigg(\varphi(x-\ell)+\varphi(x+\ell) \bigg) \bigg(\ell^3 -\ell (7 \sigma^2+5 x^2 )\bigg) + 16 \ell (2 \sigma^2+x^2 ) \varphi(x)
      \\
    &\hspace{3cm} - \bigg(\varphi(x+\ell) - \varphi(x-\ell) \bigg) \bigg(\ell^2 x +3 x (5 \sigma^2+x^2) \bigg)   \bigg\rbrack \\
    &\qqqq + \big\lbrack \ell^4-6 \ell^2 (\sigma^2+x^2 )+8 \ell (3 \sigma^2 x+x^3 )-3 (3 \sigma^4+6 \sigma^2 x^2+x^4 ) \big\rbrack \operatorname{erf}\bigg(\frac{\ell-x}{s}\bigg) \\
    &\qqqq +\big\lbrack \ell^4-6 \ell^2 (\sigma^2+x^2 )-8 \ell (3 \sigma^2 x+x^3 )-3 (3 \sigma^4+6 \sigma^2 x^2+x^4 ) \big\rbrack \operatorname{erf}\bigg(\frac{\ell+x}{s}\bigg) \\
    &\qqqq + 16 \ell x (3 \sigma^2+x^2 ) \operatorname{erf}\bigg(\frac{x}{s}\bigg) \Bigg\rbrack .
  \end{split} \label{eq:Kp_wendland2_gauss}
\end{align}
Again, similar but more complex expressions exist for higher-order Wendland kernels.
We have not found or been able to compute the integrals of the mean embeddings.

]

\clearpage

\subsection{Fractional Brownian Motion Kernels}

Let $d = 1$ and $\Omega = \lbrack a, b \rbrack$ for $b > a > 0$.
The \emph{Brownian motion kernel} is $K^{1/2}(x, y) = \min\{x, y\}$.
Let $H \in (0, 1)$ be a parameter known as Hurst index.
The Brownian motion kernel is obtained by setting $H = 1/2$ in the family of \emph{fractional Brownian motion kernels}
\begin{equation}
    K^H(x, y) = \frac{1}{2} \big( \lvert x \rvert^{2H} + \lvert y \rvert^{2H} - \lvert x - y \rvert^{2H} \big) .
\end{equation}

\textbf{Uniform Distribution ($d=1$).} Consider the uniform distribution in~\eqref{eq:lebesgue-measure} on $\Omega = \lbrack a, b \rbrack$.
Set $h=2H+1$.
Then
\begin{align}
    K_P^H(x) &= \frac{b^h - a^h-(b-x)^h - (x - a)^h}{2h(b-a)} + \frac{x^{h-1}}{2}, \label{eq:Kp_fbm_uniform} \\
    K_{PP}^H  &= \frac{(h+1)(b^h - a^h) - (b-a)^{h}}{h(h+1)(b-a)} . \label{eq:Kpp_fbm_uniform}
\end{align}
These are obtained via straightforward integration.
Note that the above corresponds to a Brownian motion with zero boundary at $x=0$, but a version with zero boundary at $x=1$ is sometimes also used in the QMC literature. See Section 2.4 of \citep{Dick2010} for details and the expressions of $K_P$ and $K_{PP}$.

\subsection{Power Series Kernels}
Let $c_\alpha \in \R$. A \emph{power series kernel} has the form
\begin{equation}
    K(x, y) = \sum\nolimits_{\alpha \in \N_0^d} c_\alpha x^\alpha y^\alpha \quad \text{ for } \quad x, y \in \R^d,
\end{equation}
where the sum is over $d$-dimensional non-negative multi-indices, operations on which are defined in the usual way.
Setting $c_\alpha = 1$ for $\lvert \alpha \rvert = 1$ and $c_\alpha = 0$ otherwise gives the linear kernel $K(x, y) = \langle x, y \rangle_2 = x^\T y$.

\vspace{0.1cm}
\textbf{Uniform Distribution.}
Consider the $d$-dimensional uniform distribution in~\eqref{eq:lebesgue-measure}.
By integrating polynomials we get
\begin{align}
    K_P(x) &= \sum\nolimits_{\alpha \in \N_0^d} x^\alpha c_\alpha \prod_{i=1}^d \frac{b_i^{\alpha_i+1} - a_i^{\alpha_i+1}}{(\alpha_i + 1)(b_i - a_i)}, \label{eq:Kp_powerseries_uniform} \\
  K_{PP} &= \sum\nolimits_{\alpha \in \N_0^d} c_\alpha \bigg(  \prod_{i=1}^d \frac{b_i^{\alpha_i+1} - a_i^{\alpha_i+1}}{(\alpha_i + 1)(b_i - a_i)} \bigg)^2 . \label{eq:Kpp_powerseries_uniform}
\end{align}

\textbf{Gaussian Distribution (Diagonal).}
Consider the centered Gaussian distribution in~\eqref{eq:gaussian-measure} with covariance $\Sigma = \mathrm{diag}(\sigma_1^2, \ldots, \sigma_d^2)$.
Since $\Sigma$ is diagonal, the formula for central moments of a univariate Gaussian yields
\begin{align}
    K_P(x) &= \sum\nolimits_{\alpha \in 2\N_0^d } x^\alpha c_\alpha \prod_{i=1}^d \sigma_i^{\alpha_i} (\alpha_i - 1)!!, \label{eq:Kp_powerseries_gaussian} \\
    K_{PP} &= \sum\nolimits_{\alpha \in 2\N_0^d } c_\alpha \bigg( \prod_{i=1}^d \sigma_i^{\alpha_i} (\alpha_i - 1)!! \bigg)^2 , \label{eq:Kpp_powerseries_gaussian}
\end{align}
where $2\N_0^d$ denotes the set of multi-indices with even elements and $n!! = 1 \cdot 3 \cdots (n-2) n$.
Isserlis' theorem could be used to compute $K_P$ and $K_{PP}$ for general $\Sigma$.

\subsection{Stationary Kernels on Spheres} \label{sec:spheres}

Let $\mathbb{S}^d = \{ x \in \R^{d + 1} : \lVert x \rVert_2 = 1\}$ denote the $d$-dimensional unit sphere and let $P$ be the uniform spherical measure on $\mathbb{S}^d$.
In this case many stationary kernels have constant embeddings.
For example, the kernel
\begin{equation}
    K(x, y) = 2 - \lVert x - y \rVert_2 \quad \text{ for } \quad x, y \in \mathbb{S}^2,
\end{equation}
whose RKHS is a Sobolev space of order $3/2$ on $\mathbb{S}^2$, has
\begin{equation}
    K_P(x) = K_{PP} = \frac{2}{3} \quad \text{ for all } \quad x \in \mathbb{S}^2 . \label{kp_kpp_sobolev32_sphere}
\end{equation}
The infinitely smooth kernel 
\begin{equation}
    K(x, y) = 48 \exp(-12\lVert x - y \rVert_2) \quad \text{ for } \quad x, y \in \mathbb{S}^2,
\end{equation}
has
\begin{equation}
    K_P(x) = K_{PP} = 1 - \exp(-48) \quad \text{ for all } \quad x \in \mathbb{S}^2 . \label{kp_kpp_sobolevinfty_sphere}
\end{equation}
See \citet{Graf2013} and \citet{Ehler2017} for these and other kernels on closed manifolds.

\emph{Periodic Sobolev kernels} of order $2r$ ($r \in \N$) are another class of kernels with constant embeddings.
They are
\begin{align}\label{eq:periodic_sobolev_kernel}
    K^{2r}(x, y) &= 1 + (-1)^{r+1} (2\pi)^{2r} \frac{\mathrm{B}_{2r}( \lvert x - y \rvert )}{(2r)!} \\
    \label{eq:korobov-expansion}
     & = 1 + 2 \sum_{k = 1}^\infty k^{-2r} \cos \big(2\pi k(x-y)\big).
\end{align}
for $x, y \in [0, 1]$, where $\mathrm{B}_{2r}$ is the Bernoulli polynomial of degree $2r$; see \citep[Ch.\@~2]{Wahba1990} for the series expansion.
Since $K^{2r}(\cdot, y)$ is continuous and periodic for each $y \in [0, 1]$, $K^{2r}$ can be viewed as a kernel on $\mathbb{S}^1$.
All terms in the sum~\eqref{eq:korobov-expansion} integrate zero.
Therefore
\begin{equation} \label{eq:korobov-embeddings}
    K_P^{2r}(x) = K_{PP}^{2r} = 1 \quad \text{ for all } \quad x \in [0, 1],
\end{equation}
where $P$ can be interpreted as either the uniform distribution on $[0, 1]$ or the spherical measure on $\mathbb{S}^1$.
See \citet{Rathinavel2019} and \citet{Belhadji2019} for uses of these kernels in Bayesian/kernel quadrature.

The identities~\eqref{eq:korobov-embeddings} remain true if  $k^{-2r}$ in~\eqref{eq:korobov-expansion} are replaced with any positive coefficients for which the series converges.
Additional terms that integrate to zero can be included without altering the embeddings in~\eqref{eq:korobov-embeddings}.
For example, the term $1$ in~\eqref{eq:periodic_sobolev_kernel} could be replaced with $\sum_{\tau=0}^r \mathrm{B}_\tau(x) \mathrm{B}_\tau(y) / (\tau!)^2$ as in \citet[Sec.\@~15.4]{Dick2010} since $\mathrm{B}_\tau$ for $\tau \geq 1$ integrate to zero.
Related kernels with constant embeddings include digital shift invariant and scramble invariant kernels~\citep[Thms.\@~12.7 \& 13.20]{Dick2010}.

\section{Building Tractable Embeddings}
\label{sec:construction}

We now discuss what to do when the pair of kernel $K$ and distribution $P$ of interest is not one of those above.

\subsection{Building on Known Kernel Embeddings}
A first approach is to obtain tractable expressions from transformations of known expressions. This trick has long been used in the literature, and was formalised by \cite{Li2021}  in the context of probabilistic circuits.

\textbf{Product Kernels and Product Distributions.}
Suppose that our distribution and kernel factorise so that
\begin{align}
    p(x)=\prod\nolimits_j p_j(x_{j}), \quad K(x,y)=\prod\nolimits_j K^j(x_j,y_j),
\end{align}
and assume that the embeddings of $P_j$ with $K^j$ are known in closed-form for all $j$. Then, the kernel embedding and its integral are given by
\begin{align}
    K_P(x) = \prod\nolimits_j K^j_{P_j}(x), \quad K_{PP} = \prod\nolimits_j K^j_{P_j P_j}.
\end{align}
Note that this approach can be straightforwardly generalised to the case of products of multivariate marginals.

\textbf{Sum Kernels and Mixture Distributions.}
Suppose that we have a mixture distribution and/or a sum kernel:
\begin{align}
    p(x)&=\sum\nolimits_j w_j p_j(x), \\
    K(x,y)&=\sum\nolimits_{j'} \gamma_{j'} K^{j'}(x,y),
\end{align}
where $K^{j'}_{P_j}$ and $K^{j'}_{P_j P_j}$ are known in closed-form for all $j,j'$.
Then, the kernel mean embedding and its integral are themselves the sum of known quantities:
\begin{align}
    K_P(x) &= \sum\nolimits_{j,j'} w_j \gamma_{j'} K^{j'}_{P_j}(x), \\
    K_{PP} &= \sum\nolimits_{j,j'} w_j w_{j'} \gamma_{j''} K^{j''}_{P_j P_j}.
\end{align}
\textbf{Change of Measure.}
Suppose we want to compute the integral $I(f)$ of a function $f$ against $P$. If $K_P$ and $K_{PP}$ are intractable, but we have access to closed-forms for $K_Q$ and $K_{QQ}$ for another distribution $Q$, then one approach is the \emph{``change of measure trick''} (or \emph{``importance sampling trick''}; \citealp[e.g.,][Sec.\@~5]{Karvonen2019}). Suppose that $P$ and $Q$ have densities $p$ and $q$.
Then
\begin{equation}
\begin{split}
  I(f)  = \int_\Omega f(x) p(x) \dif x &= \int_\Omega \left(\frac{f(x) p(x)}{q(x)}\right) q(x) \dif x \\
  & = \int_\Omega g(x) q(x) \dif x.
\end{split}
\end{equation}
This trick works for Bayesian quadrature, but cannot necessarily be used more broadly since it is not an approach for computing unknown kernel mean embeddings.

\textbf{Change of Variable.} 
Suppose that $Q$ is some distribution on $\Omega_Q$ for which we have closed-form expressions of $K_Q$ and $K_{QQ}$. If we are interested in having closed-form embeddings for the distribution $P = \varphi_\# Q$, the pushforward of $Q$ through the invertible map $\varphi\colon\Omega_{Q} \rightarrow \Omega$, then one approach is to use the \emph{``change of variable trick''} (also sometimes called the \emph{``inverse transform trick''}). This consists of using a kernel of the form
\begin{align}
    K^\varphi(x,y) = K(\varphi(x),\varphi(y)),
\end{align}
since a change of variables gives us 
\begin{equation}
\begin{split}
    K^\varphi_P(x)  & = \int_\Omega K(\varphi(x),\varphi(y)) \dif P(y) \\
    &= \int_{\Omega_Q} K(\varphi(x),y) \dif Q(y)  = K_Q(\varphi(x)).
\end{split}
\end{equation}
One simple example is to take $\varphi$ to be the inverse cumulative distribution function for $P$, in which case one can use a kernel $K$ with closed-form embeddings against the uniform distribution. This trick is particularly natural when computing embeddings with respect to simulators/generative models~\citep{Bharti2023}.

\textbf{Matrix-Valued Kernels.} In applications such as multi-output Bayesian quadrature \citep{Xi2018MultiOutput,Gessner2019,Karvonen2019,Sun2021} one works with vector-valued RKHSs \citep{Alvarez2011Review}. This leads to matrix-valued kernels $K \colon \Omega \times \Omega \to \mathbb{R}^{T \times T}$,  $T \in \mathbb{N}$. In these settings, it is often possible to recover embeddings through the embeddings of scalar-valued kernels. For example, a common construction is to take $K(x,y)=B K^{\text{s}}(x,y)$ where $K^{\text{s}}\colon \Omega \times \Omega \rightarrow \mathbb{R}$ is a scalar-valued kernel and $B \in \mathbb{R}^{T \times T}$ a positive semi-definite matrix. In this case, both the kernel mean embedding and its integral are matrices that can be directly obtained as $K_P(x) = B K^{\text{s}}_P(x)$ and $K_{PP} = B K^{\text{s}}_{PP}$.

\subsection{Stein Reproducing Kernels}

Since $K_P$ and $K_{PP}$ are known for few kernel/distribution pairs,
an alternative is to \emph{design} a reproducing kernel such that these quantities are available in closed-form.
This is the idea behind Stein reproducing kernels~\citep{Oates2017, Anastasiou2021}.
One example is the \emph{Langevin Stein reproducing kernel}, $\tilde{K}$. Suppose that $\Omega = \mathbb{R}^d$, $K$ is a sufficiently regular and $P$ satisfies $\int_{\Omega} \|\nabla_x \log p(x)\|_2 \dif P(x) < \infty$. Then
\begin{equation}
\begin{split}
    \tilde{K}(x,y) 
     ={}& K(x,y) (\nabla_x \log p(x)^\T \nabla_y \log p(y)) \\
    & + \nabla_x K(x,y)^\T \nabla_y \log p(y) \\
    & + \nabla_y  K(x,y)^\T \nabla_x \log p(x)  \\
    & + \text{Tr}(\nabla_x \nabla_y K(x,y)) .
\end{split}
\end{equation}
The $i$th entry of $\nabla_x K(x,y) \in \mathbb{R}^d$ is $\partial K(x,y)/\partial x_i$ and the $(i,j)$th entry of the matrix $\nabla_x \nabla_y K(x,y) \in \mathbb{R}^{d \times d}$ is $\partial^2 K(x,y)/\partial x_i \partial y_j$.
The kernel $\tilde{K}$ has the key property
\begin{align}\label{eq:stein_embeddings}
    \tilde{K}_P(x) = \tilde{K}_{PP} = 0 \quad \text{ for all } \quad x \in \Omega
\end{align}
by construction. Alternatively, if having a kernel mean embedding of zero is not convenient, one can use a kernel of the form $\tilde{K}^C(x,y)=\tilde{K}(x,y)+C$ for $C \in \mathbb{R}$, so that $\tilde{K}^C_P(x) = C$ and $K^C_{PP}=C$. This is particularly useful for Bayesian quadrature, in which case $C$ can be viewed as a kernel hyperparameter to be optimised.

One of the main advantages of the Langevin Stein kernel is that it only requires knowledge of $P$ through evaluations of the score function $\nabla_x \log p$, which is available for most densities known up to normalisation constant. Indeed, suppose that $p(x)=\tilde{p}(x)/Z$ where $\tilde{p}$ can be evaluated pointwise but $Z>0$ is unknown.
Then $\nabla_x \log p(x) = \nabla_x \log \tilde{p}(x)$. The score can hence be obtained through automatic differentiation using  $\tilde{p}$. This allows the user to obtain closed-form embeddings for Bayesian posterior distributions, or unnormalised models such as large graphical models and deep energy models.

The idea of using Stein reproducing kernels to obtain closed-form kernel mean embeddings has been used successfully in a broad range of application areas. It has been used for Bayesian/kernel quadrature and related control variate approaches \citep{Oates2017,Oates2016CF2,Barp2018,Karvonen2018,Si2020,Sun2021,South2022}, but also in the context of the maximum mean discrepancy, in which case the discrepancy is called \emph{kernel Stein discrepancy}. This has led to new kernel herding algorithms \citep{Chen2018,Chen2019}, gradient flows \citep{Korba2021}, goodness-of-fit tests \citep{Chwialkowski2016,Liu2016}, parameter estimators \citep{Barp2019}, and generalised posterior distributions \citep{Matsubara2021} for which kernel embeddings are available by construction.

\section{Library}
\label{sec:lib}

Kernel mean embeddings, while useful, are cumbersome to implement and test, which raises the bar for their practical usefulness. 
A small number of existing software packages contain closed-form expressions. 
These include the \texttt{ProbNum} \citep{Wenger2021} and \texttt{Emukit} \citep{emukit2023} packages in Python, and the \texttt{regMMD} package in R \citep{regMMD}. 
However, the main focus of these libraries is to provide the user with the final method (which uses a kernel embedding) rather than to make the closed-form expression itself accessible. Thus, code related to kernel embeddings is often ``hidden away'' or linked with code for other purposes. To accompany the collection of embeddings in this paper, we therefore provide an accessible Python library called \texttt{kernel\_embedding\_dictionary}\footnote{\sloppy\url{https://github.com/mmahsereci/kernel_embedding_dictionary}} whose purpose is to collect and make available embeddings in one place. 

The structure is simple: i) we first instantiate a kernel mean embedding object and ii) then evaluate it at $x$. The example below shows how to do this for the uniform distribution (Lebesgue measure) with a Gaussian kernel.

\begin{mdframed}
\begin{lstlisting}[language=Python,
basicstyle=\scriptsize,
]
from kernel_embedding_dictionary import \
    get_embedding

ke = get_embedding("expquad", "lebesgue")

# evaluate kernel mean embedding
x = np.random.randn(3, 1)
ke.mean(x)
\end{lstlisting}    
\end{mdframed}
\vspace{-3mm}
\begin{mdframed}[backgroundcolor=lightgray2]
\begin{lstlisting}[language=Python,
basicstyle=\scriptsize,
%basicstyle=\ttfamily\small,
%commentstyle=\color{blue}\sffamily\textit
]
carray([0.92829 , 0.844137, 0.334471])
\end{lstlisting}    
\end{mdframed}

Parameters of the distribution and the kernel can be defined in a configuration. 

\begin{mdframed}
\begin{lstlisting}[language=Python,
basicstyle=\scriptsize,
]
config_measure = {
    "ndim": 2,
    "bounds": [(0, 1), (-1, 0.5)],
    "normalize": True
}

config_kernel = {
    "ndim": 2,
    "lengthscales": [1.0, 2.0],
}

ke = get_embedding("expquad", "lebesgue", \
    config_kernel, config_measure)
\end{lstlisting}    
\end{mdframed}

All our embeddings are unit tested and neatly listed in the form of functions in a single module. This makes it easy to find and reuse the appropriate code for the user's own scientific project (MIT License). Given its intended use, the library does not provide any methods that use kernel embeddings and is not optimized for efficiency. The library can be thought of as a dictionary of kernel embeddings ``in the form of code,'' providing one of the cumbersome building blocks for writing more elaborate project code. 
We hope that over time \texttt{kernel\_embedding\_dictionary} will become a point of reference and contain a representative collection of embeddings contributed by the open source community. 

\section{Conclusion}

This paper provides a dictionary of kernel/distribution pairs for which the kernel mean embedding and its integral have a known closed-form, and reviewed several approaches to construct new expressions. Our hope is that this will save many a researcher the time needed to derive or implement kernel embeddings from scratch.

Additional related integrals not discussed in this paper are also occasionally needed. For example, some extensions of Bayesian quadrature~\citep{Gunter2014,Deisenroth2009, Pruher2018} require integrating certain products of kernels not covered in this paper. Expanding our dictionary with these expressions could therefore be useful. Some algorithms also require embeddings of conditional distributions \citep[Sec.\@~4]{Muandet2016}; see for instance \citep{Chen2024conditionalbq} for their use in Bayesian quadrature. Several of the embeddings above can already be intepreted as embeddings of conditional distributions, but again expanding our dictionary with this focus in mind could be of interest.
  
Of course, the paper would be incomplete without mentioning the broad literature studying approximations of kernel mean embeddings. 
In principle, any quadrature rule can be used~\citep{Sommariva2006}.
Given independent samples $x_1,\ldots,x_n$ from $P$, the most common approximation is obtained through a Monte Carlo estimator: $K_P(x) \approx \frac{1}{n} \sum_{i=1}^n K(x,x_i)$, which \citet{Tolstikhin2016} show to be minimax-optimal and for which finite-sample bounds are available in \citet{Wolfer2022}. Several other estimators have also been proposed, including a shrinkage estimator \citep{Muandet2015}, a kernel density estimation-based estimator \citep{Sriperumbudur2016}, a Gaussian process-based approach \citep{Flaxman2016}, a quasi-Monte Carlo estimator \citep{Niu2021} and even a Bayesian quadrature estimator \citep{Bharti2023}.
These can typically improve the error rate or provide uncertainty quantification, but at the cost of additional regularity assumptions.
In certain cases, approximating embeddings is easier than in others.
For example, it may be known that the embeddings are constant (as in Section~\ref{sec:spheres}), so that only one integral needs to be approximated, or there may be symmetries that drastically reduce the number of approximations  needed~\citep{Karvonen2018, Karvonen2019}.

\subsubsection*{Acknowledgements}
The authors would like to thank  Siu Lun Chau, Zonghao Chen, Krikamol Muandet, Masha Naslidnyk, Luc Pronzato, and Anatoly Zhigljavsky for helpful discussions. 
Equation~\eqref{eq:Kpp_matern_general_uniform} is due to Anatoly Zhigljavsky. FXB was supported by the EPSRC grant [EP/Y022300/1].
TK was supported by the Research Council of Finland grant 359183.

\appendix

\bibliography{references}

\end{document}